\documentclass{article}

\usepackage{arxiv}
\usepackage{amsmath}

\usepackage[utf8]{inputenc} 
\usepackage[T1]{fontenc}    
\usepackage{hyperref}       
\usepackage{url}            
\usepackage{booktabs}       
\usepackage{amsfonts}       
\usepackage{nicefrac}       
\usepackage{microtype}      
\usepackage{lipsum}
\usepackage{graphicx}
\graphicspath{ {./images/} }
\usepackage{caption}

\title{Towards Reinforcement Learning Based Log Loading Automation}

\author{
  Ilya Kurinov \\
  School of Energy Systems\\
  LUT University\\
  Yliopistonkatu 34, 53850 Lappeenranta, Finland \\
  \texttt{ilya.kurinov@student.lut.fi} \\
  \And
  Miroslav Ivanov \\
  School of Arctic Natural Resources and Economy\\
  Lapland University of Applied Sciences\\
  Jokiväylä 11, 96300 Rovaniemi, Finland \\
  \texttt{miivanov@edu.lapinamk.fi} \\
  \And
  Grzegorz Orzechowski \\
  School of Energy Systems\\
  LUT University\\
  Yliopistonkatu 34, 53850 Lappeenranta, Finland \\
  \texttt{grzegorz.orzechowski@lut.fi} \\
  \And
  Aki Mikkola \\
  School of Energy Systems\\
  LUT University\\
  Yliopistonkatu 34, 53850 Lappeenranta, Finland \\
  \texttt{aki.mikkola@lut.fi} \\
}

\begin{document}
\maketitle

\begin{abstract}
Forestry forwarders play a central role in mechanized timber harvesting by picking up and moving logs from the felling site to a processing area or a secondary transport vehicle. Forwarder operation is challenging and physically and mentally exhausting for the operator who must control the machine in remote areas for prolonged periods of time. Therefore, even partial automation of the process may reduce stress on the operator. This study focuses on continuing previous research efforts in application of reinforcement learning agents in automating log handling process, extending the task from grasping which was studied in previous research to full log loading operation. The resulting agent will be capable to automate a full loading procedure from locating and grappling to transporting and delivering the log to a forestry forwarder bed. To train the agent, a trailer type forestry forwarder simulation model in NVIDIA's Isaac Gym and a virtual environment for a typical log loading scenario were developed. With reinforcement learning agents and a curriculum learning approach, the trained agent may be a stepping stone towards application of reinforcement learning agents in automation of the forestry forwarder. The agent learnt grasping a log in a random position from grapple's random position and transport it to the bed with 94\% success rate of the best performing agent.
\end{abstract}


\section{Introduction}
Forest forwarders are heavy-duty machines perform tasks such as lifting, transferring, and positioning logs in complex forest environments. They combine articulated hydraulic cranes and grippers, which provide the necessary strength and dexterity required to manipulate logs of various sizes and weights with precision and stability~\cite{shevchuk}. Designed for rugged outdoor conditions, forest forwarders must perform reliably on rough terrain and through obstructions such as trees, bushes, or other equipment.

Currently, human operators are responsible for the control of these machines. They operate from raised cabins with clear views, or they control the equipment remotely using teleoperation technologies. The control interfaces of modern forwarders may include pedals and joysticks or state of the art haptic or assistive feedback devices for precise control of the crane and grapple~\cite{hap}. The operation of a forest forwarder is physically demanding, requiring operators to maintain a high level of focus and precision continuously for extended periods of time~\cite{stress}. This is due to the combination of continuous fine motor control, repetitive movements, and the need for sustained concentration over long shifts. Operators must manipulate joysticks, pedals, and control panels with precision, often requiring  small frequent adjustments to the crane’s hydraulic system. Maneuvering massive logs with a machine that has multiple degrees of freedom demands excellent spatial awareness, quick decision-making, and fine motor skills. The stress associated with crane operation is amplified by the potential risk of accidents, equipment malfunctions, and pressure to meet productivity targets. 

As the mechanical complexity and operating demands of log handling increase, interest in automating forwarders using Artificial Intelligence (AI) and robotics is growing. Autonomous or semi-autonomous operation of such machines not only holds the promise of improving operational efficiency and lowering operator fatigue, but also of improving safety by limiting human exposure to hazardous environments. Among various approaches, reinforcement learning is particularly well-suited for training intelligent agents to execute control policies in complex systems such as forest forwarders via training in simulated environments.

This research aims to continue exploration of the application of reinforcement learning algorithms in forestry forwarder automation, focusing on automating full cycle of log picking. The previous studies discussed in literature review focus mainly on automating forwarders only till grabbing stage or using conventional methods. Grabbing of the log is a challenging task, but continuing the motion from grabbing is a substantially more complex task. Additional complexity of the log loading from grabbing motion comes from longer horizon of the task and shifting task objectives, such as grabbing, transporting and loading to the bed. 

The key contributions of this paper are:
\begin{enumerate}
    \item The study extends previous works in forestry forwarder automation. The aim is to develop a reinforcement learning agent which is capable not only grasping but lifting, transporting and loading log to the bed.
    \item Demonstrating the feasibility of the reinforcement learning agent for tasks with a long horizon and identifying potential issues in the approach.
    \item Defining the example action and observation spaces with suitable reward function.
    \item Identifying benefits and pitfalls of the approach. Identify a potential way of deploying the agent in the real-world application.
\end{enumerate}

\section{Literature review}
The topic of the forestry forwarder automation appears in multiple studies over the years. Studies, relevant to the topic of this research and  related to the automation using artificial intelligence, fall into categories of application of reinforcement learning and convolutional neural networks. Studies applying reinforcement learning methods focus on automating only grasping motion. In the research by \cite{andersson} studied automation of the grasping motion using reinforcement learning with curriculum on a simulation model of an experimental forestry forwarder. The study applied a curriculum with distance from initial position increase during the training and an energy optimization objective. The best agent was able to grab logs with 97\% accuracy, while energy optimization model is able to grasp logs with 93\% success rate. The researchers propose for the future studies to focus on application of the agent in more unstructured environment.  Another study by \cite{wallin} applies reinforcement learning and virtual visual servoing for multi-log grasping.  The study uses reinforcement learning agent which takes an image from a virtual camera (with CNN for identifying target grasping position) and trained on curriculum with energy optimization objective. The resulting agent is able to grasp logs with 95\% overall success rate, 97\% for three logs grasping and 91\% for five logs grasping. In the study by \cite{dhakarte} investigated automation of the log grasping process using convolutional neural network. The study trained an agent in the simulated environment and futher tested on a scaled forwarder crane in a laboratory environment. Two agents were trained to control the crane to follow the predefined path tracks. The study by \cite{wiberg} automates forwarder's locomotion task using reinforcement learning. The study applies a forestry forewarder model and scans of a real harvesting site terrain for reconstructing the environment. The resulting agent is capable of traversing steep angle terrain and applying different strategies for locomotion.  

There are studies which are focusing on forestry forwarder operation using other methods than reinforcement learning. The research by \cite{ayoub} studied application of CNN for grasp planning of a log. The grasp planning pipeline uses input from depth camera for recreating the scene in Gazebo simulation environment with creating grasp map. The tests were conducted on a real forestry machinery and proved to have a robust grasping with maximum success rate of 98.3\%. Another study by \cite{semberg} shows application of a single stereo camera for log detection and grasping. The system was able to recognize the graple, the log and it's gasping point with further calculation of boom motion vector. The resulting system was tested on a real forestry forwarder with grasp success of 82\%. The research by \cite{lahera} developed a system for autonomous operation of the forestry forwarder. The system is capable of navigating the job site, locating the log, loading to he bed and delivering to the final destination. Mentioned capabilities were achieved by combining multiple methods in one system: the locomotion uses GPS and predefines path tracking point of a mission, the log detection uses depth camera and DNN for tracking and estimating position of the logm the log handling was Dynamic Movement Primitives algorithm, which is trained on demonstrations from experienced operators. The systems was capable of fully autonomous operation in repeatable and stable way.

\section{Methods}
Reinforcement Learning is a type of machine learning and optimal control approach that addresses sequential decision-making problems. An agent learns to make decisions by interacting with an environment to maximize cumulative rewards over time ~\cite{suttonReinforcementLearningIntroduction2018}. Unlike other forms of learning, RL focuses on actions that have delayed rewards. The agent explores the environment, taking actions that lead to new states and receiving rewards. The goal is to learn a policy that maps states to actions to achieve the highest long-term reward. A key challenge in RL is balancing exploration (trying new actions to discover their effects) and exploitation (choosing actions known to yield high rewards). 

Reinforcement learning is widely used for game testing~\cite{rlGameTesting}, robotics~\cite{rlRobotics}, autonomous vehicles~\cite{rlAV}, healthcare~\cite{rlHealthcare}, and finance~\cite{rlFinance}.
It is applied in various ways in mechanical engineering and heavy machinery control. For example, it is used to optimize the operation of robotic arms in manufacturing, which ensures precise and efficient movements ~\cite{huangRoboticArmVelocity2023}. RL also helps in the autonomous control of heavy machinery such as excavators and bulldozers, allowing them to perform complex tasks such as digging, loading, and material handling with minimal human intervention~\cite{kurinovAutomatedExcavatorBased2020, huhDeepLearningBasedAutonomous2023}. These applications demonstrate RL’s ability to handle complex decision-making problems with delayed rewards, significantly improving efficiency and performance.

Formally, an RL task is often modeled as a Markov decision process defined by a tuple data structure $ (S, A, P, R, \gamma) $, where $ S $ is the state space, $ A $ is the action space, $ P(s'|s,a) $ is the state transition probability, $ R(s,a) $ is the reward function, and $ 0 \leq \gamma < 1 $ is a discount factor weighting future rewards~\cite{suttonReinforcementLearningIntroduction2018}. The agent’s behavior is governed by a \textit{policy} $ \pi(a \mid s) $ that specifies the choice of action in each state. The RL algorithm determines how the policy responds to environmental interaction.  

The quality of a policy can be evaluated by \textit{value functions}. The state-value $ V(s) $ gives the expected total discounted reward from state $ s $ (following the policy thereafter). These functions satisfy recursive Bellman equations; in particular, the Bellman optimality equation for the optimal value function is: 

\begin{equation}
    V^*(s) = \max_{a} \mathbb{E} \left[ R(s,a) + \gamma V^*(s') \right],
\end{equation}
where $ s $ is a state, $a$ is an action, $R(s,a)$ is a reward for a state $s$ and an action $a$, $\gamma$ is a discount factor and $V^*(s')$ is a value funstion for the next state. 

Equation can be used to find the optimal value of a state equals the maximum, over all actions, of the expected sum of the immediate reward and the discounted optimal value of the next state. Multiple different RL algorithms are developed to learn such optimal values. Common methods include advantage actor-critic~\cite{mnih_asynchronous_2016}, PPO~\cite{schulmanProximalPolicyOptimization2017}, soft actor-critic~\cite{haarnoja_soft_2018}, and deep Q-networks~\cite{mnih_playing_2013}.

\subsection{Curriculum Learning for Reinforcement Learning}
\label{subsec: curriculum reinforcement learning}

Curriculum Learning (CL) is a training strategy where an agent learns through a sequence of tasks with increasing difficulty~\cite{bengio_curriculum_2009}. In reinforcement learning, it is used to improve learning efficiency, especially in problems with sparse rewards and long-horizon tasks. Rather than training an agent on the full task from the start, CL initially introduces simplified versions of the task and gradually increases complexity~\cite{bengio_curriculum_2009}.

In sparse reward environments, agents often fail to reach the goal via random exploration. CL addresses this by structuring tasks so that agents first learn sub-skills with intermediate rewards~\cite{narvekar_curriculum_2020}. For example, in robotic manipulation, agents might first learn to reach for an object, then grasp it, and finally place it at a target location~\cite{florensa_reverse_2017}. This method improves credit assignment and accelerates learning.

CL is also used for tasks comprising multiple phases. By dividing the task into stages, each with its own reward, the agent can focus on one subtask at a time. In this project, for example, PPO was used to train a forest forwarder to perform a sequence of actions: locate a log, grasp, lift, place on a trailer, and return to the initial position. A simple curriculum was applied by assigning rewards to these sub-tasks, which guided the learning process and reduced the burden of exploration. See Section~\ref{subsubsec: reward function} for more details.

Various strategies exist to design curricula, including manually defined stages and automated methods~\cite{florensa_reverse_2017}. Studies show that curriculum learning improves sample efficiency and leads to higher-quality policies. For robotic tasks specifically, it has been effective in enabling agents to learn complex behaviors with fewer training steps.

\subsection{Proximal Policy Optimization}

PPO is a popular reinforcement learning algorithm that balances simplicity and performance. It is a type of policy gradient method that optimizes the policy directly by updating it to avoid large, disruptive changes. It achieves this by limiting the policy change using a clipping mechanism, which helps maintain stability during training~\cite{schulmanProximalPolicyOptimization2017}. This makes PPO more robust and easier to tune compared to other algorithms such as trust region policy optimization.

In mechanical engineering, particularly in heavy machinery control, PPO is highly effective due to its ability to handle complex, continuous control tasks. For example, PPO can be used to optimize the operation of autonomous excavators or bulldozers~\cite{kurinovAutomatedExcavatorBased2020}. By learning the most efficient and safe way to perform tasks such as digging, loading, or material handling, PPO can improve precision and efficiency. The algorithm’s robustness makes it suitable for environments where stability and safety are critical, which ensures that machinery can operate autonomously with minimal human intervention while adapting to changing conditions and tasks. This leads to more efficient operations, reduced wear and tear on equipment, and improved overall productivity.

\subsection{Isaac Gym}
Isaac Gym is state-of-the-art physics simulation software developed by NVIDIA. It was designed to train and RL agents in virtual environments that accurately represent real-world scenarios. The Isaac Gym applies the parallelization capabilities of consumer-grade Graphics Processing Units (GPUs) to simultaneously simulate thousands of environments, a capability that posed significant challenges for conventional simulators and required a cloud or a datacenter. Simultaneously simulating thousands of environments makes possible significantly more data collection and policy learning, which is particularly useful in areas such as robotics, where real-time interactions with the environment are computationally expensive and time-consuming.

One of the main characteristics of Isaac Gym is close coupling with high-speed rigid-body simulation and RL frameworks based on PyTorch, which allows for large-scale training on the GPU alone. This design reduces the need for data movement between GPU and central processing unit memory in the training cycles, which reduces latency and enhances throughput. The simulator sustains articulated and rigid-body systems with complex multibody dynamics support, contact simulation, and collision handling, features critical to simulating the real-world robotic system behaviors of, for example, manipulators, legged robots, or forestry equipment such as forest forwarders. Additionally, Isaac Gym provides domain randomization capabilities and configural flexibility, which enables sim-to-real transfer and robust policy learning in a variety of dissimilar environments.

Both from a research and engineering perspective, Isaac Gym offers a powerful platform for the study of scalable RL algorithms, curriculum learning strategies, and scenarios relevant to real-world deployment. Its ability to run thousands of simulation environments in parallel enables~\cite{makoviychuk_isaac_2021} the study of sample efficiency, convergence dynamics, and generalization across varied task variations. Furthermore, the simulator's modularity enables the use of custom reward functions, multi-modal sensing, and a variety of control devices, which renders it a flexible platform for experimentation on a broad range of robotics and control problems. Isaac Gym illustrates a fundamental building block in designing next-generation autonomous systems trained using deep RL methods.

\section{Problem statement}
The study applies reinforcement learning control to solve a loading task. As shown in Figure~\ref{fig:forwarder-operation}, this task involves performing multiple  nontrivial consecutive steps to load the log. As RL is typically applied to single-step problems~\cite{janner2021offline}, solving such a complicated task requires a mastering curriculum. This study introduced the involved stages by designing several separate rewards for grabbing, lifting, and dropping a log. The environment was designed and implemented in Isaac Gym~\cite{makoviychuk_isaac_2021}, and control was implemented via PPO agent. 

\subsection{Forestry forwarder operation} \label{operation}
Forestry forwarder operation is a sequence of tasks designed to transport harvested logs from the harvesting area to the landing area or a road from where the collected logs are transported for further processing. The forwarder operation cycle consists of traveling to the cutting area, loading logs, moving them to the piling site, and unloading them. 

The typical forester forwarder operation sequence begins by traveling to the harvesting site. Depending on the type of forestry forwarder, it may drive or be transported to the site. An operator must consider the topology of the terrain, the type and size of the wood products, visibility, and weather conditions~\cite{donaldFwdManual2018}.

\begin{figure}[!t]
    \centering
    \includegraphics[width=\linewidth]{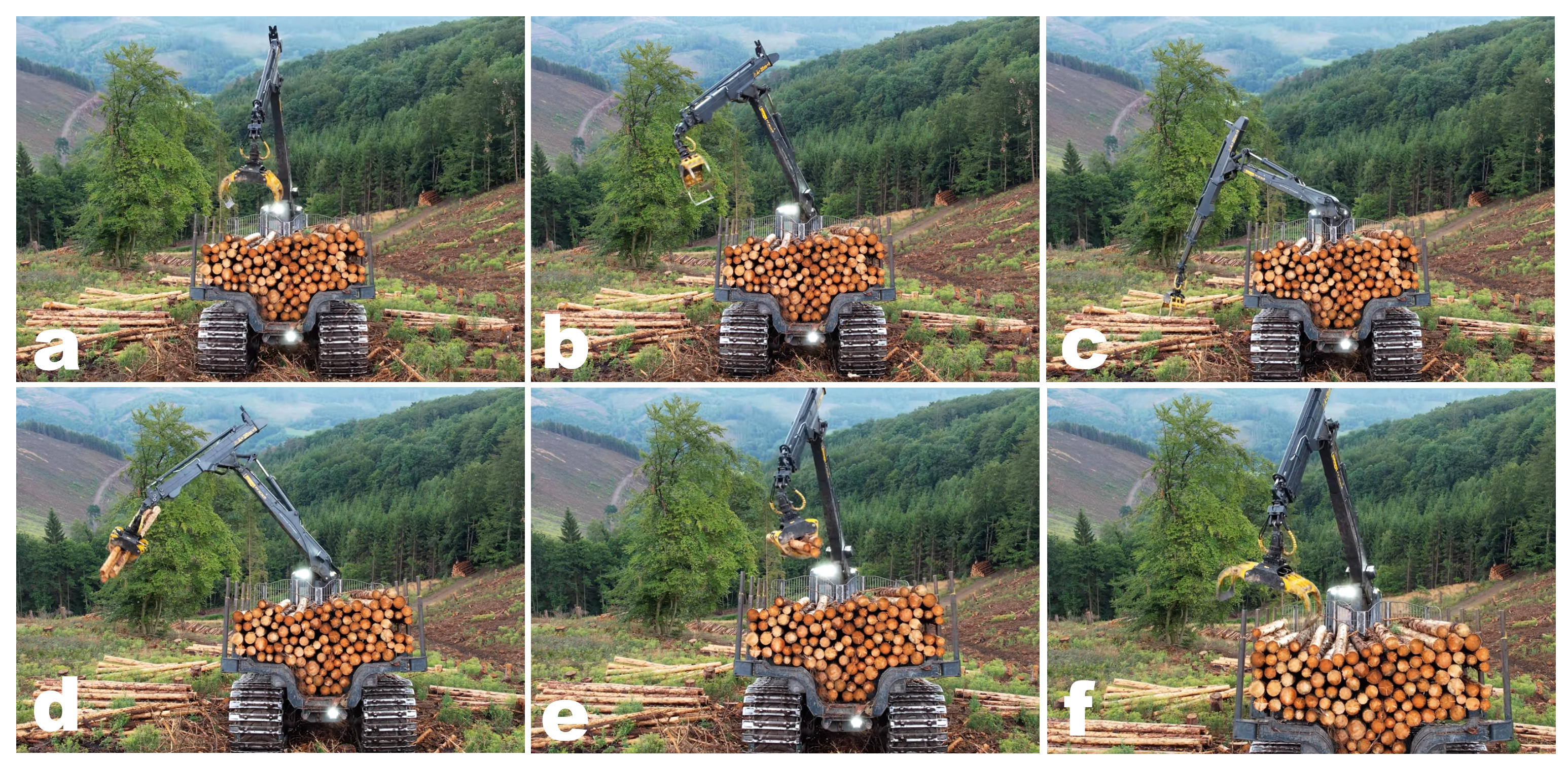}
    \setcounter{footnote}{0}
    \caption[caption]{Forwarder operation cycle\footnotemark: a) The grapple is at the initial position above the bed. b) The forwarder reaches the log pile position. c) The grapple closes around the logs. d) The crane lifts logs above the bed. e) The pile of wood is lowered onto the bed. f) The grapple releases.} 

    \label{fig:forwarder-operation}
\end{figure}
Once positioned near the log pile, the forwarder uses the hydraulic crane equipped with grapple to pick up logs. See Figure~\ref{fig:forwarder-operation}. The grapple grasps a log, lifts it above the forwarder bed, and lowers it onto the bed. Proper log arrangement is critical to optimizing load capacity and stability. This process continues until the bed is fully loaded or all harvested logs have been handled.

Once the bed has been loaded, the forwarder transports the logs from the harvesting area to a piling area where the logs are loaded onto a truck or stored for further transport by truck. Similarly to the transportation step, a forwarder operator must consider terrain topology and weather conditions, and he must account for changes in the center of mass to handle this step safely and effectively.

The entire cycle, from the trip to the cutting area, loading, transporting, unloading, and returning,repeats continuously throughout the timber harvesting process, allowing efficient flow of wood from the forest to the processing or transport sites. The productivity of the forwarder and its cycle time depend on factors such as terrain conditions, log arrangement, and the distance between the cutting area and the landing site~\cite{fwdProductivity}. Optimal operation requires skilled maneuvering, precise crane control, and fluent loading/unloading techniques, all of which contribute to the total productivity of the forestry operation.

\subsection{Forestry forwarder model description} \label{model_description}
The simulation model imitates a trailer-type forest forwarder equipped with a crane and a grapple attachment. It is an open-loop system with 9 degrees of freedom. The main parts of the machine are a four-wheel trailer used for the transportation and storing of logs and a hydraulic crane with  grapple for log handling. The list of bodies is presented in \ref{table:fwdBodies}
\begin{figure}[!t]
    \centering
    \includegraphics[width=\linewidth]{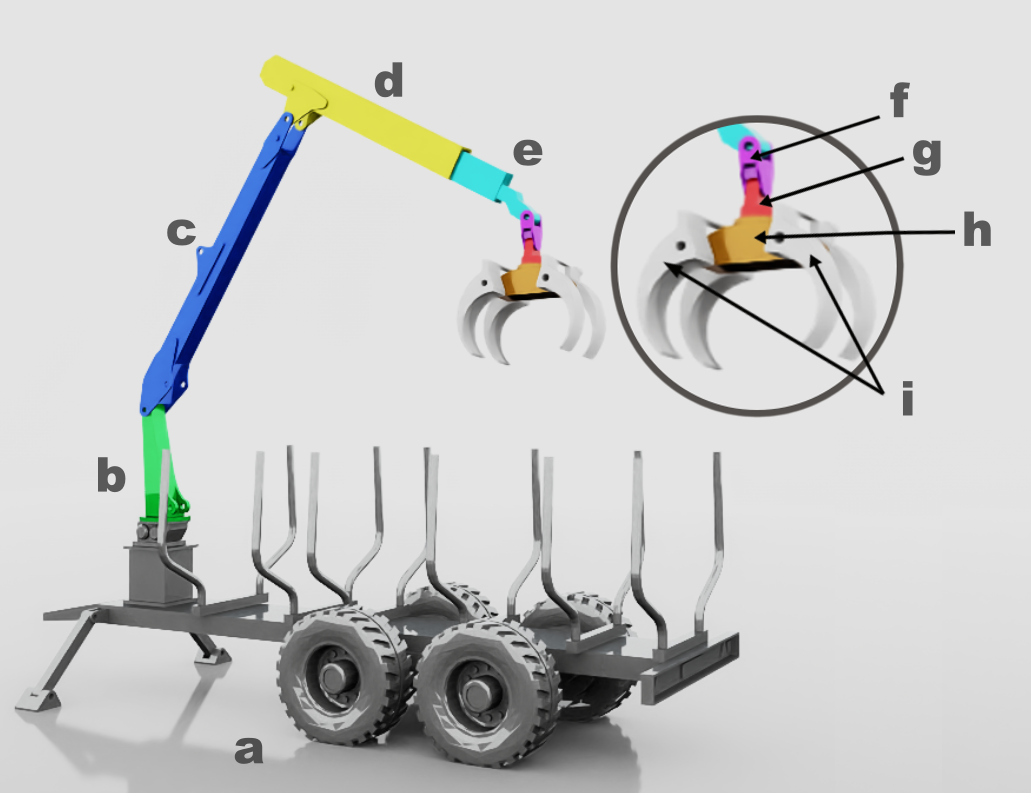}

    \caption{Model topology of forwarder: a) base with trailer and bed, b) manipulator base, c) crane arm, d) extension arm, e) extension, f) intermediate hook, g) grapple rotator, h) grapple body, i) grapples.}

    \label{fig:forwarder-model}
\end{figure}
The base of the trailer is the first body in the model, which contains the bed and four wheels. The wheels in the simulation serve a decorative function, and they are part of the base. In this simulation, therefore, no suspension model was implemented. In the simulation environment, the base is fixed to the ground plane. The collision graphics of the bed was approximated with a box shape, and therefore moving of the grapple between bed poles is restricted despite it being allowed in the real-world operation scenario.

The manipulator crane comprises five bodies. The base of the crane is attached to the trailer with a rotational joint, which is allowed to move 180 degrees, i.e., 90 degrees to each side. The crane arm is responsible for vertical reach and lifting capabilities. The crane arm has a rotational joint with a limit of 70 degrees. The extension arm is a body with an extension boom attached. With motion limited to 70 degrees, it is responsible for the reaching and, partially, lifting capabilities. The extension boom attaches to the extension arm with a slider joint, which makes possible an extension to 1.5 meters to assist in reaching a load. The intermediate hook is a body that attaches the grapple assembly to the crane. It permits rotational motion of the grapple assembly along the $x$- and $y$-axes, providing its free movement to facilitate the grabbing motion. A 210-degree rotational joint connects the grapple assembly to the crane, and a 180-degree rotational joint connects to the grapple rotator.

The grapple assembly consists of four bodies: the grapple rotator, the grapple body, and two grapples. The grapple rotator connects the grapple to the intermediate hook. The rotational joint connected to the grapple body can rotate 180 degrees to match the rotation of the grapple body holding a log. This ensures a successful grab. All of the aforementioned bodies have a simplified convex mesh, which was not altered. The grapples connect to the grapple body with a 75-degree rotational joint.  For ensuring the optimal log grab, collision meshes for the grapples were altered and are represented by two boxes, since convex meshes cannot easily replicate their shape and the convex hull decomposition method provides excessive simulation overhead cost.
\begin{table}[!t]
    \caption{Names, types and limits of the forestry forwarder model joints}
    \label{table:fwdJoints}
    \centering
    \begin{tabular}{ lllccl  }
    \hline
    Joint Name                  &Symbol &Joint type &Min limit  &Max limit  &Units    \\
    \hline
    Base -- Manipulator base          &$j_1$    & Revolute  & $-70$            & 70           & Degrees  \\
    Manipulator Base -- Crane arm     &$j_2$    & Revolute  & $-30$            & 40           & Degrees  \\
    Crane arm -- Extension arm        &$j_3$    & Revolute  & $-30$            & 40           & Degrees  \\
    Crane arm -- Extension            &$j_4$    & Slider    &  0             & 1.5          & Meters   \\
    Extension -- Intermediate hook    &$j_5$    & Revolute  & $-90$            & 120          & Degrees  \\
    Intermediate hook -- grapple rotator &$j_6$    & Revolute  & $-90$            & 90           & Degrees  \\
    Grapple rotator -- grapple body      &$j_7$    & Revolute  &  0             & 180          & Degrees  \\
    Grapple body -- Grapple left      &$j_8$    & Revolute  & $-45$            & 30           & Degrees  \\
    Grapple body -- Grapple right     &$j_9$    & Revolute  & $-45$            & 30           & Degrees  \\
    \hline
    \end{tabular}
\end{table}

\begin{table}[!t]
    \caption{Names and masses of the forestry forwarder model}

    \label{table:fwdBodies}
    \setlength{\tabcolsep}{3pt}
    \centering
    \begin{tabular}{ lc}
    \hline
    Body name         &Mass,kg \\
    \hline
    Base               &1500    \\
    Manipulator Base   &158    \\
    Crane arm          &171    \\
    Extension arm arm  &132    \\
    Extension          &75    \\
    Intermediate hook  &16    \\
    Grapple rotator       &12    \\
    Grapple body       &50   \\
    Grapple left       &40    \\
    Grapple right      &40    \\
    \hline
    \end{tabular}
\end{table}
Table \ref{table:fwdJoints} represents a detailed description of the joints and their characteristics. Defining joint limits is important to approximate the behavior of the actual forestry forwarder and accurately replicate its motion constraints. Real machines often use hydraulic cylinders and motors for actuation. Therefore, due to closed-loop structure of the crane, joint motions are limited. Because the simulation model has an open-loop structure, joint limits were introduced to imitate actual machine behavior. This makes it possible to simulate a machine with relatively accurate movements for the agent to solve a log handling task and limit the workspace to reduce the complexity of the task.

\subsection{Task}
The previously described model solves a simplification of log loading that is close to the real-world scenario. The operation involves a forestry forwarder simulation model, a ground plane and a log model. The simulation environment is also simplified. The ground plane is flat, and the log parameters do not change during simulation. The agent’s goal is to move the log from a random starting position onto the bed using the controls of the forwarder simulation model.

\subsubsection{Action space}
In the context of reinforcement learning, an action space is the complete set of possible actions that an agent can perform at any state in a particular environment. It specifies the output domain of the policy that maps states or observations into subsequent actions. Action spaces are generally classified into two categories: discrete and continuous. A discrete action space contains a finite set of different actions It is typically used in settings such as games, where the agent selects among a pre-defined set of options. In a continuous action space, the agent can select from a range of values for each action dimension. It is frequently used in applications such as robotic control, where actions involve smooth motor commands.

The action space plays a crucial role in determining complexity with regard to the learning problem. Larger or more continuous spaces result in higher-dimensional search problems for the policy that require using more advanced exploration and optimization methods. The form and shape of the action space are essential in determining the effectiveness of the agent to interact with and productively affect its environment.

The action space of the forest forwarder model constitutes joint position goals, which are incremented by a scaled action provided by the agent. Each active joint is scaled by the corresponding scaling factor and a timestep $dt$. This approach was taken to reduce the velocity of the base of the crane, the crane arm, the extension arm, and the extension. Bigger position increments result in erratic movements of the arm, unstable behavior of the grapple, and an inability to grab the log. Shorter joint increments allows more precise and safer movement. The resulting action vector is clipped to the joint limits discussed previously in the Forestry Forwarder Model Description. Equation \ref{eq:action_space} describes the action space.
\begin{center}
    \begin{equation} \label{eq:action_space}
        \mathbf{U} = \mathrm{clip}(\mathbf{q} + \mathbf{A} \cdot \mathbf{c} \cdot dt , u_{min}, u_{max})
    \end{equation}
\end{center}
where $\mathbf{U}$ is a joint position goals matrix of size $n_{joints} \times n_{envs}$, $
\mathbf{q}$ is a current joints position matrix of size $n_{joints} \times n_{envs}$, $\mathbf{A}$ is an action provided by the agent, $\mathbf{c}$ is a scaling factor vector, and $dt$ is a time step. The joint position matrix has a following form.
\begin{equation}
    \begin{aligned}
        \mathbf{U}
        &=
        \begin{bmatrix}
        u_{j11} & u_{j21} & u_{j31} & u_{j41} & u_{j71} & u_{j81} & u_{j91} \\
        \dots & \dots & \dots & \dots & \dots & \dots & \dots\\
        u_{j1n} & u_{j2n} & u_{j3n} & u_{j4n} & u_{j7n} & u_{j8n} & u_{j9n} \\
        \end{bmatrix}
        \\
    \end{aligned}    
\end{equation}
where $u_{j1n}$, $u_{j2n}$, $u_{j3n}$, $u_{j4n}$, $u_{j7n}$, $u_{j8n}$, $u_{j9n}$ are joint positions, and $n$ is a number of environments.

\subsubsection{Observation space}
The RL observation space refers to the set of all possible states or sensory inputs that an agent can perceive from the environment at any given time. It defines the structure, dimensionality, and boundaries of the data that the agent receives to make decisions. Formally, the observation space is a subset of the environment's state space, which contains complete information about the environment's current configuration.

However, the agent may only receive partial information, which constitutes its observation. The space can be continuous, as in physical attributes such as position or velocity, or discrete, as in categorical data such as different terrain types. The design and representation of the observation space are crucial, because they directly influence the agent’s ability to learn and generalize effectively from the environment. A well-defined observation space ensures that the agent can extract relevant features to make informed decisions, which affects the overall performance of the RL algorithm. The observation space of the forestry forwarder model is represented by four main elements: a forwarder parameters vector $\mathbf{S}_{fwd}$, the grapple body parameters vector $\mathbf{S}_{g}$, a log parameters vector $\mathbf{S}_{l}$, and the task parameters vector $\mathbf{S}_{task}$.

The first element of the observation vector describes the state of the forwarder simulation model. The forwarder parameters vector includes joint positions and joint velocities. Joint positions describe rotational positions of the active joints in radians. Joint velocities describe the angular velocities of the forwarder's active joints in radian/sec. These parameters provide vital spatial information for the agent, which helps to determine an optimal course of action. The resulting forwarder parameters vector has the following form.
\begin{equation}
    \begin{aligned}
        \mathbf{S}_{fwd}
        &=
        \begin{bmatrix}
        u_{j1} & \dots & u_{jm} & \dot{u_{j1}} & \dots &  \dot{u_{jm}} \\
        \end{bmatrix}
        \\
    \end{aligned}    
\end{equation}
where $p_{jm}$ and $\dot{p_{jm}}$ are joint positions and velocities of active degrees of freedom of the forwarder model, and $m$ is a joint number. The active degrees of freedom of the forwarder model are $j_1, j_2, j_3, j_4, j_7, j_8, j_9$ as described in Table \ref{table:fwdJoints}.

The second part of the observation vector describes the parameters of the grapple body. This vector provides information about the grapple body and the grapples. The vector contains parameters of the grapple body as x, y, and z position in world coordinates in meters, velocity in m/s, and rotation in quaternion form. This data helps to navigate and align the grapple body with the log. The velocity of the grapple body helps the agent decide the velocity of approach to the log or bed to avoid hard contact with the ground or machine. The left and right grapple positions provide world-coordinate data that further aid in achieving accuracy of alignment of the grapple with either the log or load bed. The grapple body observation vector has a following form.

\begin{equation}
    \begin{aligned}
        \mathbf{S}_{g}
        &=
        \begin{bmatrix}
        p_{gb} & H_{gb} & \dot{p_{gb}} & p_{gl} & p_{gr}\\
        \end{bmatrix}
        \\
    \end{aligned}    
\end{equation}
where $p_{gb}$, $p_{gl}$, $p_{gr}$ are xyz positions of the grapple body and left and right grapples in world coordinates, $\dot{p_{gb}}$ is a velocity of the grapple body, and $H_{gb}$ is the rotation of the grapple body in quaternions.

The third part of the observation vector is the log parameters. This vector shows the position in world coordinates and the orientation of the log in quaternion form. Along with grapple body parameters, the log parameters are essential for accurate grapple aiming and log transport. The resulting vector has a following appearance.

\begin{equation}
    \begin{aligned}
        \mathbf{S}_{l}
        &=
        \begin{bmatrix}
        p_{l} & H_{l}\\
        \end{bmatrix}
        \\
    \end{aligned}    
\end{equation}
where $p_{l}$ is a position vector, and $H_{l}$ is a rotation in quaternions of the log.

The last part of the observation vector describes important task parameters and degree of completion. The $p_{unl}$ and $p_{tgt}$ parameters were introduced to evaluate completion of the log handling task. The $p_{unl}$ parameter describes position of the imaginary point located above the center of the bed and bed guards. Establishing this intermediate target point on the way to the target helps to navigate without following the direct path. The target point $p_{tgt}$ signifies log target position, which is located at the bottom of the bed. Terms $p_{gb}-p_l$, $p_l-p_{unl}$, and $p_l-p_{tgt}$ are the difference in positions along the $x$-$y$-$z$ axes. They provide the agent with the relative position to the grapple body, the unloading point, and the target. The $a_l$ parameter is a variable indicating completion of the task. It gives a score from 0--1 if the log is delivered to the target point.
\begin{equation}
    \begin{aligned}
        \mathbf{S}_{task}
        &=
        \begin{bmatrix}
        p_{unl} & p_{tgt} & p_{gb}-p_{l} & p_{l}-p_{unl} & p_{l}-p_{tgt}  & a_{l}\\
        \end{bmatrix}
        \\
    \end{aligned}    
\end{equation}
where $p_{unl}$ is the position of an unloading point situated above the bed, $p_{tgt}$ is the position of a target point inside of the bed, $a_{l}$ is a variable indicating completion of the log transport, and $p_{gb}-p_{l}$, $p_{unl} - p_{l}$, and $p_{l}-p_{tgt}$ are the position differences between the grapple body to the log, the log to the unloading point, and the log to the target, respectively. The $a_l$ variable can be expressed by the following equation.
\begin{equation}
    \begin{aligned}
        a_l = x_{ltgt} \cdot x_{lgb} \cdot x_{gbtgt}
    \end{aligned}    
\end{equation}
where $x_{ltgt}$ is a score of reducing distance between the log and the target, $x_{lgb}$ is a score of reducing distance between the log and the grapple body, and $x_{gbtgt}$ is a score of reducing distance between the grapple body and the target point. These scores are calculated as follows.

\begin{equation}
    \begin{aligned}
        x = \frac{1}{1+d^2}
    \end{aligned}    
\end{equation}
where $d$ is a Euclidian distance between two positions.
Eventually, the resulting observation vector can be represented by following equation.
\begin{equation}
    \begin{aligned}
        \mathbf{S}
        &=
        \begin{bmatrix}
        \mathbf{S}_{fwd1}  & \mathbf{S}_{g1} & \mathbf{S}_{l1} & \mathbf{S}_{task1}\\
        \dots & \dots & \dots & \dots\\
        \mathbf{S}_{fwdn}  & \mathbf{S}_{gn} & \mathbf{S}_{ln} & \mathbf{S}_{taskn}\\
        \end{bmatrix}
        \\
    \end{aligned}    
\end{equation}
where $n$ is the number of training environments. The observation vector provides the required sensor reading for an agent to decide on the course of action to achieve the goal. Several parameters used in the construction of the observation vector, such as the distance reduction scores denoted by $x$, were used to formulate the reward function.

\subsubsection{Reward function}
\label{subsubsec: reward function}
The reward function is used to provide feedback from the forwarder model on actions taken. The reward given offers insight on the efficiency of reaching, grabbing, lifting, and delivery of the log. Introducing the multiplicative factor $b$ allows the agent to focus on a particular task during the training process. Therefore, the agent's behavior is shaped not only by its grabbing and lifting performance, but also by its ability to move the log towards the bed and complete the unloading phase, which is critical to maximizing cumulative reward.

The reward function has three main elements: $r_1$, $r_2$,and $r_3$. The first element r1 is responsible for guiding the grapple body towards the center of mass of the log. The $r1$ reward is especially important at the beginning of the process because the grapple is at a random position and must navigate to the log position to grab it. $r1$ takes the following form.

\begin{equation}
    \begin{aligned}
        r_1 = x_{lgb}
    \end{aligned}    
\end{equation}
where $x_{lgb}$ is the score of the distance reduction between the center of mass of the log and the grapple body. See the previous equation 8.

Part $r_2$ of the reward function is responsible for manipulating the log before loading it onto the bed of the forwarder. Scalar $r_2$ identifies movement in the $z$-axis and towards the unloading point located above the bed. Scalar $r_2$ can be described by following equation.

\begin{equation}
    \begin{aligned}
        r_2 = p_{lz} + x_{lunl} b
    \end{aligned}    
    \label{r2_reward}
\end{equation}
where $p_{lz}$ is the $z$-axis position of the log, $x_{lunl}$ is the score of the distance reduction between the log and the unloading point, and $b$ is the weighting factor. The weighting factor $b$ plays an important role in the movement sequence, making the agent focus on moving towards the unloading point.

Part $r_3$ of the reward function manages the final movement of the grapple towards the target. This part emphasizes the importance of two aspects of movement: bringing the log to the target point $p_{tgt}$ and reducing the velocity of the log in the $z$-axis. The velocity reduction is critical, since it prevents the agent from dropping or hitting the log hard, which could lead to equipment damage. The scaling factor $b$ presented in equation \ref{r2_reward} is scaled to the power of 3, emphasizing the importance of the “current” over the “previous”. $r_3$ can be expressed with the following equation.

\begin{equation}
    \begin{aligned}
        r_3 = \frac{x_{ltgt}^2}{\dot{p_{lz}}} b^3
    \end{aligned}    
\end{equation}
where $x_{ltgt}$ is the score for reducing the distance between the log and the target point, $\dot p_{lz}$ is the velocity of the log in the $z$-axis, and $b$ is the scaling factor.

The resulting reward function depends on the chosen hierarchical learning configuration. Rewards can be arranged in four ways. The first requires $r_1$, $r_2$, and $r_3$ to be a separate stage of the hierarchical learning, adding subsequent reward after completing the stage. The second and third combine two of the reward elements into one stage. The fourth option combines all of the rewards into one stage. The most promising training combination is $r_1r_2+r_3$. This implies that the model is pretrained on a reward function that consists of the sum of $r1$ and $r_2$.  The second half of the training process continues onward from the best model from the previous step.

\section{Results}

Subsequent paragraphs describe an agent trained on two stages of curriculum training. The reward for the first training stage is the sum of $r_1$ and $r_2$. The second stage's reward is the sum of stage one reward and $r_3$.

The resulting agent was able to learn the optimal policy to operate the forwarder. It adapts to environment changes such as random positions of the forwarder joints and random position of the log. It is capable of aligning the orientation of the grapple body to the log, enabling log grasping. In cases where the log is outside of the pick-up script, such as behind the bed or places out of reach, the agent struggles to get a hold. In rare cases, the agent may push the log out of reach.

In general, however, the agent is capable of moving the log and positioning it without touching the side guards of the bed. Depending on the grasping position of the log, the agent may touch the bed guards, a permitted behavior that poses no damage risk to the machine because no direct contact between any other machine components is involved. 

\begin{figure}[!t]
    \centering
    \includegraphics[width=\linewidth]{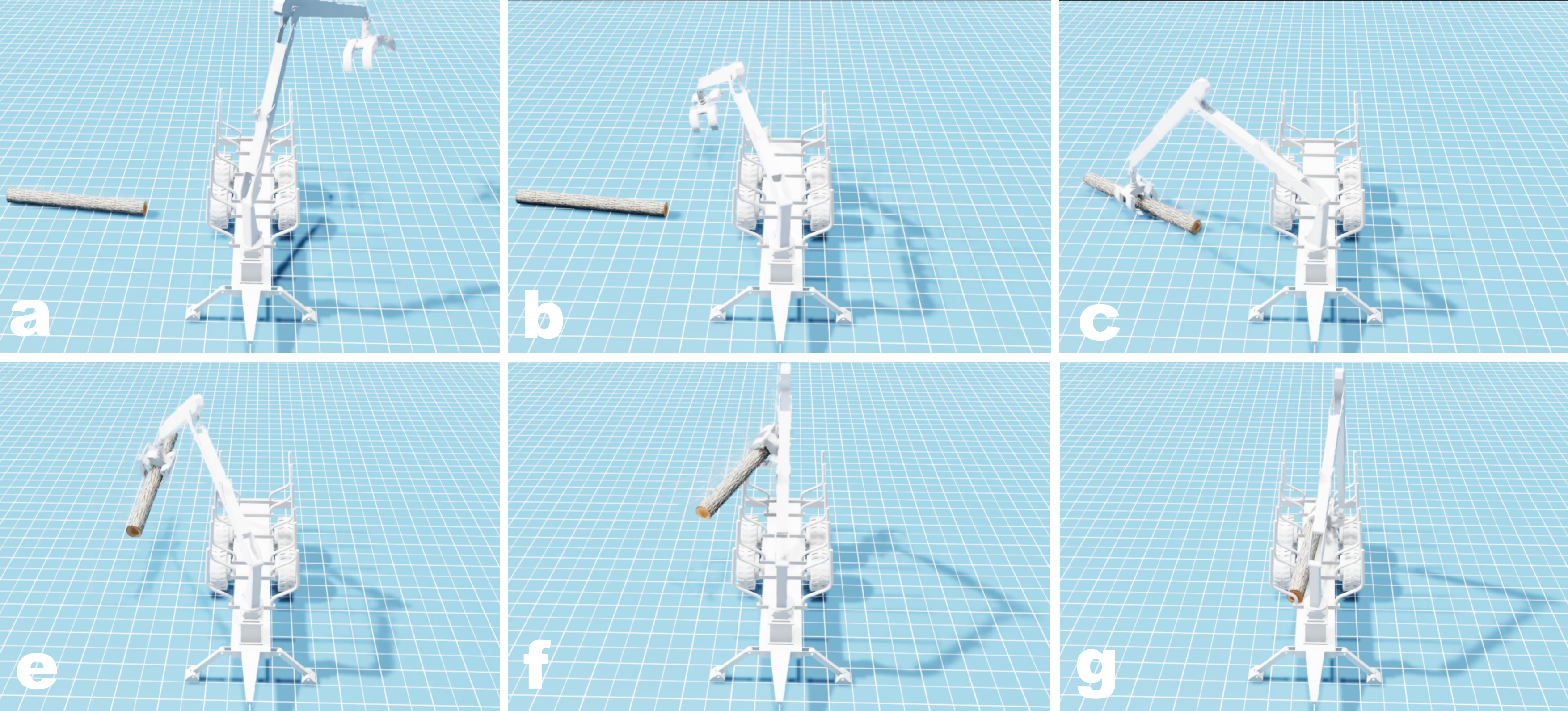}

    \caption{Result of training: The Forest forwarder agent successfully delivers the log to the bed. a) The agent starts with random joint and log positions. b) The agent navigates the grapple to the log. c) The grapple attachment reaches and grasps the log. d) The agent lifts the log and begins to move to the unloading point position. e) The log is transferred to the unloading position. f) The log is delivered to the target point located inside the bed. g) The manipulator places the log into the bed.}

    \label{fig:motion_figure}
\end{figure}
Figure \ref{fig:motion_figure} shows a sequential visual representation of the learned behavior of the reinforcement learning agent for executing the log picking task with a simulated forestry forwarder in Isaac Gym. The sequence illustrates the successful execution of the task, decomposing it into a number of key subtasks executed by the agent, which demonstrates the success of the learned policy to integrate multiple-degree-of-freedom arm control with object manipulation.

In the first example \ref{fig:motion_figure}.a, the forwarder is presented in its default initial state with the manipulator arm retracted and a target log set on the ground in front of the vehicle.  In \ref{fig:motion_figure}.b, the agent begins by extending the manipulator arm towards the log. The motion reflects that the policy has adapted to localizing the object and planning a collision-free path to reach it. The next frame \ref{fig:motion_figure}.c depicts the gripper over the log, with fine adjustments being performed by the manipulator in order to acquire a good orientation for grasping. This stage demands good coordination among arm articulation and end-effector placement.

After alignment with the log, the agent performs a grasping action from \ref{fig:motion_figure}.c to \ref{fig:motion_figure}.e, followed by a lifting stage. As indicated by \ref{fig:motion_figure}.e, the log has been successfully grasped and lifted off the floor, which indicates that the agent has learned not just to reach out and grasp but also to ensure grip stability. Frame \ref{fig:motion_figure}.f illustrates the transport phase that lifts the log over the cargo space of the forwarder. Finally, as illustrated in \ref{fig:motion_figure}.g, the manipulator places the log into the bed.

This completes the entire process of perception, approach, manipulation, and placement, and these actions demonstrate the agent's ability to navigate the environment and relocate the object without any collisions or loss of grip. Successful execution of this sequence proves that the policy employs a coherent and generalizable strategy to complete a complex, multi-step log loading task. The success confirms the reward shaping and curriculum learning methods implemented during training and demonstrates their contribution to policy generalization and stability. The fact that the agent can successfully do this in a series of iterations also suggests that the learned policy is robust to log position variation and environmental change.

\subsection{Reward}

\begin{figure}[!t]
    \centering
    \includegraphics[width=\linewidth]{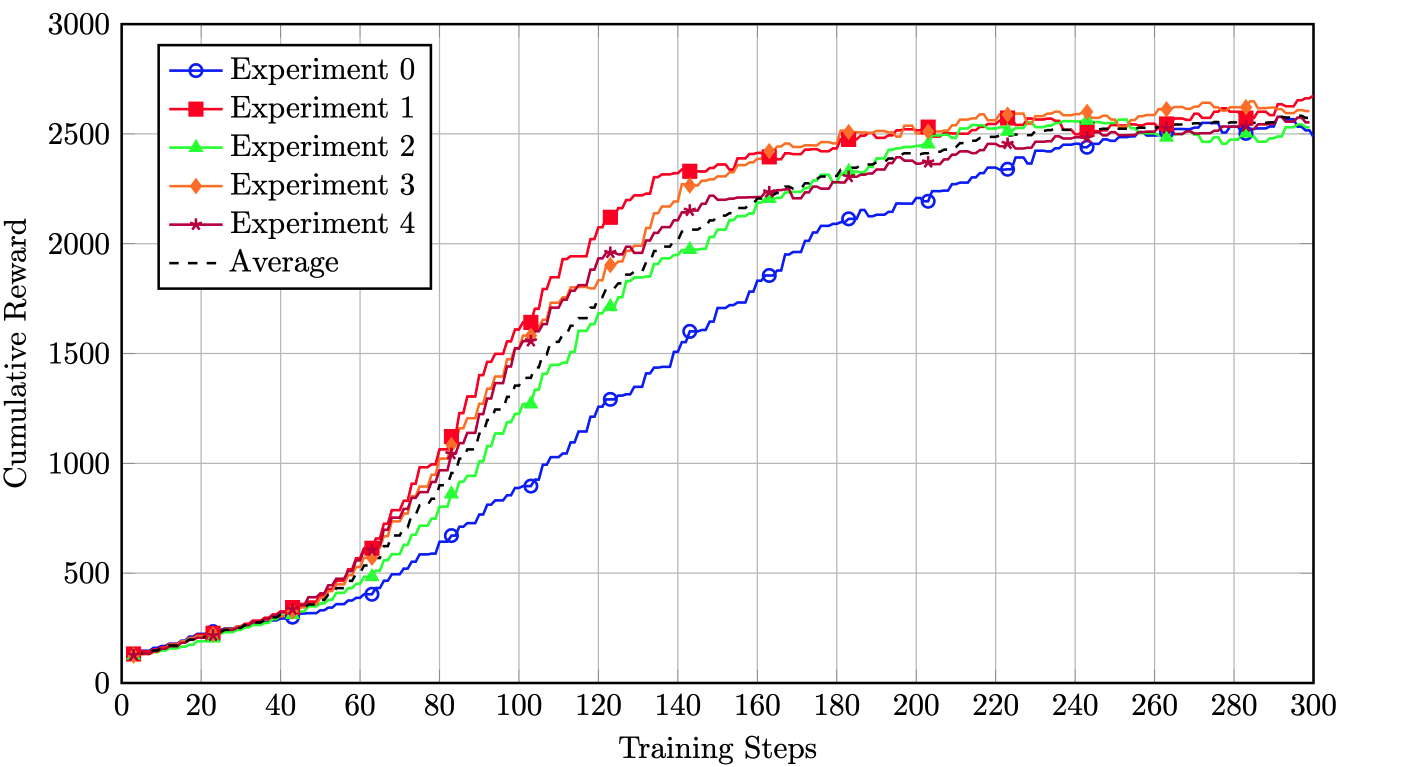}

    \caption{Step one reward: Comparison of cumulative reward over training steps for different experiments including the average curve.}

    \label{fig:reward_1}
\end{figure}

Figure \ref{fig:reward_1} illustrates the progression of cumulative rewards over the course of the first phase of curriculum learning, where the agent is trained only on using the main reward component. This phase plays a vital role. The intention is to allow the agent to establish core skills such as grasping, lifting, and moving logs into proximity and above the bed. The plot presents the results of five separate training trials (Experiments 0–4). The dashed line shows the average performance over all trials. All early trials exhibit minimal reward accumulation; however, they consistently increase as training continues, suggesting effective skill development.

There is noticeable difference in learning rate and final performance across the experiments. For instance, experiments 1 and 3 achieve higher cumulative rewards sooner, converging towards approximately 120–140 training steps, which indicates quicker convergence. Experiment 0 displays a more gradual increase, achieving similar performance only after reaching approximately 220 steps. This kind of inconsistency is common in deep reinforcement learning, and it is attributed to the randomness present in policy updates, initializations, and exploration mechanisms. Nonetheless, all experimental outcomes converge to the same cumulative rewards in the end, therefore illustrating the robustness and consistency of the learning process in this initial phase.

The average reward curve (the dashed line) smooths out oscillations and gives a good overview of the overall learning trends. It features a steep rise between the 60 and the 140 step markers, marking the crucial period during which the agent evolves from random or useless behaviors to purposive control. After 160 steps, the curve levels off, reflecting that the agent has maximized the learning potential for the first-stage reward. This leveling off validates the curriculum design. After the basic skill set is learned by the agent, the reward can be fine-tuned to move the log into the bed.

\begin{figure}[!t]
    \centering
    \includegraphics[width=\linewidth]{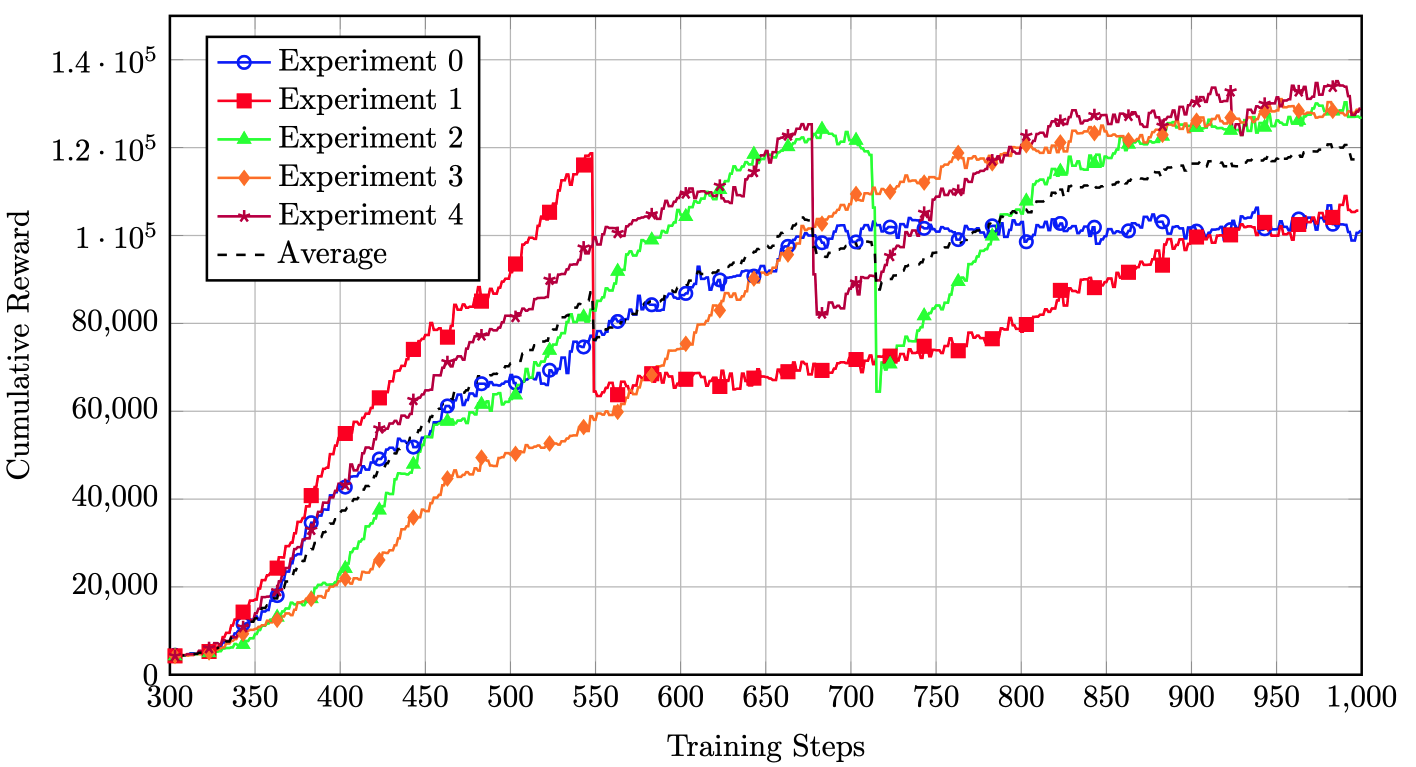}

    \caption{Step two reward: Comparison of cumulative reward over training steps for different experiments including the average curve.}

    \label{fig:reward_2}
\end{figure}
Figure 5 illustrates the cumulative reward trajectories for the second half of the curriculum learning process, i.e., with regards to the training duration from 300 to 1000 steps. During this phase, the reward function is augmented with the introduction of a novel component that promotes vertical motion of the log and good unloading behavior. Extending the understanding abilities learned in phase one, this phase teaches the agent to carry out more intricate manipulations such as avoiding collision with bed's guard and safely loading the log into the bed.

The figure highlights significant differences in how various experiments adapt to the newly established reward goal. Experiment 2 has the fastest performance increase, with a cumulative reward of over $1.2 \cdot 10^5$ at step 700, although it suffers from a sudden drop caused by policy instability or exploration noise. Experiment 3 shows the same trend, with good early learning and maintained performance. Experiments 0 and 1 demonstrate more modest but steady growth. The variation between runs captures the difficulty of scaling task complexity. The agent must recall not only previously acquired habits, but it also must modify them to facilitate successful log lifting and unloading under a novel reward regime.

The average reward curve, indicated by the dashed line, reflects the general trend observed across all runs. It shows steady and significant progress with diminishing returns after step 800. This trend suggests that the agents benefit from a curriculum of rewards that add complexity incrementally so that behavior can be refined step-by-step without overwhelming the learning process. This changeover between stages would appear to be successful since most agents utilize the first successful log handling policy to ramp to task completion that entails vertical positioning. The results confirm that curriculum learning enables more seamless adaptation to new reward objectives by promoting behavioral continuity and policy reuse.

\subsection{Evaluation}

\begin{figure}[!t]
    \centering
    \includegraphics[width=0.8\linewidth]{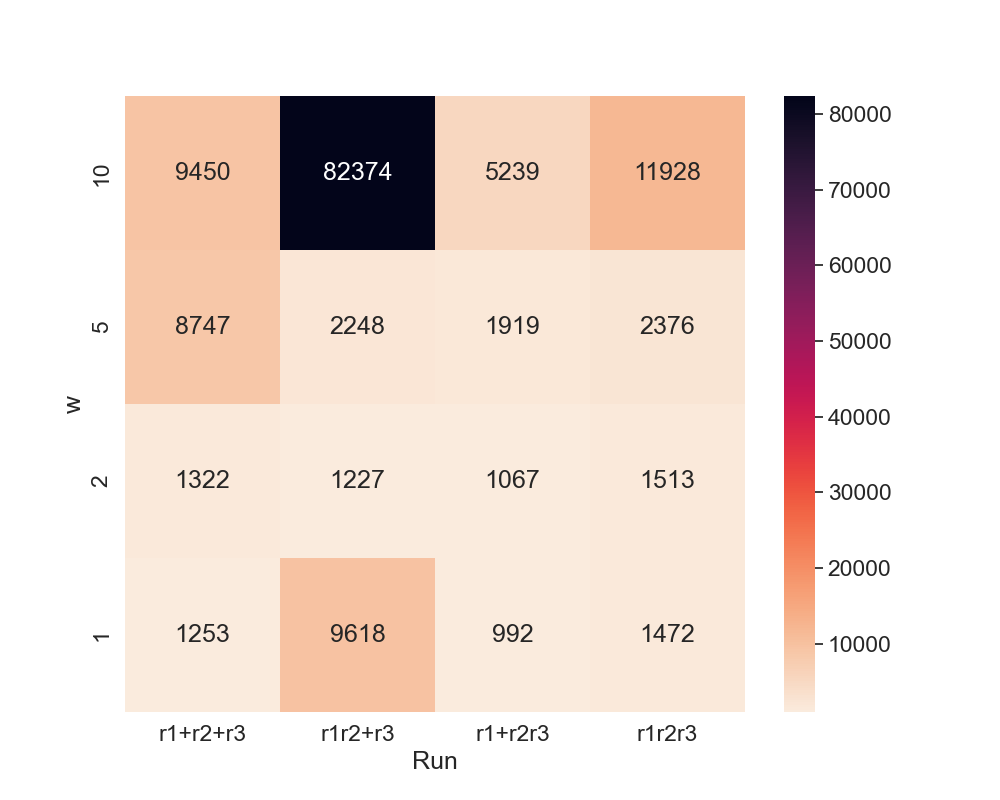}

    \caption{Returns for various versions of curriculum: Results after 600 combined epochs}

    \label{fig:experiments_heatmap}
\end{figure}

Figure \ref{fig:experiments_heatmap} is a heatmap of the cumulative reward earned by agents with four curriculum configurations. These configurations demonstrate methods of composition for the reward functions with each including three distinct components: $r1$, $r2$, and $r3$. Specifically, $r1$ incentivizes successful location of the log, r2 incentivizes lifting and reaching the unloading point above the bed, and r3 incentivizes reaching the bottom of the bed with motion stability. It accomplishes this by penalizing vertical log velocity with strongly weighted impact through cubic weighting.

The most rewarding environment is $r1r2+r3$ with weighting factor w equal to 10. This two-stage curriculum first trains the agent using a combination of grasp, lift, and moving to the point above the bed rewards so that it can learn robust manipulation skills. Once stable behavior has been learned, the full reward with the term focused on finishing the unloading r3 is added in the second stage to fine-tune. The increased final reward in this setup illustrates the value of a progressive reward curriculum. Basic abilities are acquired prior to the introduction of higher-level refinement goals to avoid early penalization or gratification during initial exploration.

The arrangement labeled $r1+r2+r3$, in which the agent learns in three sequential phases beginning with $r1$, followed by $r2$, and concluding with $r3$, results in reduced cumulative rewards. This implies that learning in excessively separated phases can discourage the agent from developing cohesive strategies that encompass several objectives. For example, training to reach the log without the simultaneous context of grasping (or vice versa) could generate poor motor patterns that fail to generalize when all the elements are later reunited.

The $r1r2r3$ configuration, in which the agent receives the complete compound reward from the start, performs worse than stage-wise curricula. This outcome demonstrates the challenge of optimizing an extremely nonlinear reward structure from the ground up. The large effect of the stability term $r3$, cubically scaled and rewarding stable vertical movement, can dominate early training stages and lead the learning process off course prior to building underlying control policies. Without a preliminary stage of organized supervision, the agent might not be able to discover purposeful behavior sequences.

Additionally, in all the explored configurations, an increase in the reward scaling factor w always results in better cumulative rewards. This indicates that stronger reinforcement signals enhance the effect of well-organized reward elements, especially in curriculum-based learning. However, the gap between curriculum-based training and flat training (e.g., $r1r2r3$) remains considerable even with heavier weights, demonstrating the inherent strengths of progressive learning methods.

The success rate of the best-performing agent with 1024 parallel environments in the Isaac Gym simulator was measured as part of the final evaluation. Success in an episode was defined as the agent successfully performing the entire log loading task: pickup, lift, and placing down the log appropriately within environment time constraints. The agent accomplished 966 out of 1,024 trials, which is equivalent to a 94\% success rate. This superior performance demonstrates the stability of the policy acquired under a wide variety of initial conditions, which validates the curriculum-guided reward shaping approach used during training.

For evaluatuion of the generalization abilities of the agent, series of tests were performed. The tests were designed to test the agent's response to unseen states in the training. The tests involved elevated ground planes, variations of wood sizes, inclinations of the forestry forwarder and rough terrain.

\section{Discussion}
The most of trained agents discussed in evaluation chapter are capable of transferring the log, but the difference comes from performance and ocurrence of failure behaviors. The most common failure behaviors, which attribute to decrease in success rate, are associated with grasping behavior, when the agent is not capable of grasping the log. The one failures with grasping often comes from the forwarder trying to reach the log that located at the positions close to the ends of the loading ares, when grapple approaches the log from the side, leading to pushing the log to the area off-limits for the arm. Some of the agents learnt a physics exloitative behavior, when the log been pushed (not lifted) to the side of the machine and using the collision to lift log by pushing it up the machine side.

To check the generalization capabilities of the best agent, number of experiments were conducted with parameters never used in the training. These experiment show that the agent is capable of partial generalization and in future works, requires domain randomization in height elevation and ranging terrain elevation. The first experiment test the ability of the agent to handle the log of the changing size. The first experiment condusts evaluation of performance with a small change ($\pm10\%$) in log radius. The experiment shows a small degradation in the performance of the best agent of to 80\% success rate with the smaller radius logs, whereas it shows no degradation or slightly better performance. In experiments with higher scale difference of $\pm25$--$50\%$, it shown a higher performance degradation, which might be due to the collision failure (grapples penetrating through the log resulting in losing it) or with smaller logs not securely attached (the joint limit not allowing to hold the log tightly). 

The best agent's performance for the experiments are  as follows: $-50\%$ scale -- $18\%$,  $-25\%$ scale -- $30\%$, $+25\%$ scale -- $78\%$, $+50\%$ scale -- $51\%$. The higher success rate in experiments with higher log scale may indicate, that most of the failure points attribute to physics and simulation settings. After adjusting the joint limits to accomodate a log with higher scale, the agents shown smaller degradation compared to previous experiment, with the best agent achieving shown $80\%$ success rate with $-10\%$ scale and no degradation with the rest of the log scales.  The another experiment shows how the agent performs the log handling with elevated ground, where the log is placed on an elevated plane. The experiment places a plane above the ground with step size of 0.2m up to 1m height. The experiments show, that with each increase of elevation, the agent loses around 13\% in performance. The final test evaluated a capabilities of the agent to handle the log with an uneven terrain that was simulated with closesly positioned cubes with ranging height eleveation. The experiment shows, that the best performing agent is capable to handle logs with 76\% rate. 

The possibility of transferring of the agent to the real-world machine can be judged by feasibility of obtaining of required parameters. The most of the parameters used for creating actions, observation and rewards consits of positions of the log and parts of the machine. The positions of the log center, grapple body and grapples may be obtained with stereo cameras or lidar, combined with a segmentation or object detection  \cite{semberg}. The joints positions and velocities may be collected by a joint position sensors and accelerometers. The rest of the parameters in reward or in observation are derrived from the positions. The sim-to-real transfer may start from further training of the agent on a high fidelity simulation model with hydraulics. After training and validating the agent on high fidelity simulation model, transfer may be done with virtual environment presented in \cite{ayoub}, where log and surrounding environment is recreated in the virtual environment. The virtual environment will allow agent may have several tries, before manipulating the real machine and decide the best performing action sequence.

\section{Conclusion}
The study resulted in an autonomous forestry forwarder trained to handle logs within the Isaac Gym simulation environment. The best performing agent is capable to locating, grasping, transporting and loading the log to the bed with 94\% success rate. The integral part of the approach used is curriculum and reward shaping that decomposes the reward into three components: moving towards the log ($r_1$), lifting and moving it to the unloading point above the bed ($r_2$), and unloading it with stability constraints ($r_3$). Each of these components corresponded to a subtask of the log handling process. Introducing each progressively through curriculum learning helped reduce the complexity of the task for the agent. This structured reward design simplified the learning problem by guiding the agent through manageable subtasks and focusing it on one objective at a time.

The experiments demonstrate that curriculum-based sequencing of reward components improves training stability and final performance. Agents trained with the progressive curriculum achieved higher success rates in handling logs than those trained with the full composite reward from the outset. The phased introduction of objectives enables the agent to master each subtask sequentially, which reduces training oscillations and accelerates convergence. The findings confirm that curriculum learning is an effective strategy to decompose complex tasks and optimize reinforcement learning outcomes. The best performing resulting agent is able to locate, grasp, lift and transport the log to the bed. The study shown a potential set of parameters, action and observation spaces for automating the log loading process.

Even though experiments were showing a possibility of the automation of the forestry forwarder with reinforcement learning, the study has a number of factors that limit the end performance of the agent. Despite these limitations, the agent showing an ability partial generalization, showing a performance degradation in unseen scenarios. Solving the limiting factors of the training process may result in higher generalization capabilities and give possibility for smoother transfer to the real-world application. The study applies an old version of the Omniverse Isaac Gym, which has a limited capability for randomization of the environment and lacks simulation capabilities of hydraulics, therefore, the agent was trained in a simplified environment. Therefore, for the future studies suggested continuation of training or using the current agent on a higher fidelity simulation model, e.g. in a Mevea simulator. One of the areas, which future studies may explore is using multiple agents for each of the tasks, swithing after successfull completion of the previous tasks.

\bibliographystyle{apalike}
\bibliography{sample}

\begin{thebibliography}{}

\bibitem[Agrawal and Mitra, 2024]{rlHealthcare}
Agrawal, S. and Mitra, P. (2024).
\newblock {\em Deep Reinforcement Learning in Healthcare and Biomedical Research}, pages 179--205.

\bibitem[Andersson et~al., 2021]{andersson}
Andersson, J., Bodin, K., Lindmark, D.~M., Servin, M., and Wallin, E. (2021).
\newblock Reinforcement learning control of a forestry crane manipulator.
\newblock {\em CoRR}, abs/2103.02315.

\bibitem[Ayoub et~al., 2023]{ayoub}
Ayoub, E., Levesque, P., and Sharf, I. (2023).
\newblock Grasp planning with cnn for log-loading forestry machine.
\newblock In {\em 2023 IEEE International Conference on Robotics and Automation (ICRA)}, pages 11802--11808.

\bibitem[Bai et~al., 2025]{rlFinance}
Bai, Y., Gao, Y., Wan, R., Zhang, S., and Song, R. (2025).
\newblock A review of reinforcement learning in financial applications.
\newblock {\em Annual Review of Statistics and Its Application}, 12(Volume 12, 2025):209--232.

\bibitem[Bengio et~al., 2009]{bengio_curriculum_2009}
Bengio, Y., Louradour, J., Collobert, R., and Weston, J. (2009).
\newblock Curriculum learning.
\newblock In {\em Proceedings of the 26th {Annual} {International} {Conference} on {Machine} {Learning}}, pages 41--48, Montreal Quebec Canada. ACM.

\bibitem[Chen, 2025]{rlRobotics}
Chen, Z. (2025).
\newblock Research on autonomous navigation and control of unmanned surface vehicles based on reinforcement learning algorithms and multi-objective optimization models.
\newblock In {\em Proceedings of the 4th International Conference on Computer, Artificial Intelligence and Control Engineering}, CAICE '25, page 365–372, New York, NY, USA. Association for Computing Machinery.

\bibitem[Dhakate et~al., 2022]{dhakarte}
Dhakate, R., Brommer, C., Bohm, C., Gietler, H., Weiss, S., and Steinbrener, J. (2022).
\newblock Autonomous control of redundant hydraulic manipulator using reinforcement learning with action feedback.
\newblock In {\em 2022 IEEE/RSJ International Conference on Intelligent Robots and Systems (IROS)}, pages 7036--7043.

\bibitem[Donald et~al., 2018]{donaldFwdManual2018}
Donald, K., Boswell, B., Amishev, D., and Hunt, J. (2018).
\newblock {\em WINCH-ASSIST FORWARDER:BEST PRACTICE MANUAL}.
\newblock FPlnnovations, Pointe-Claire, QC, Canada.

\bibitem[Florensa et~al., 2017]{florensa_reverse_2017}
Florensa, C., Held, D., Wulfmeier, M., Zhang, M., and Abbeel, P. (2017).
\newblock Reverse {Curriculum} {Generation} for {Reinforcement} {Learning}.
\newblock In {\em Proceedings of the 1st {Annual} {Conference} on {Robot} {Learning}}, pages 482--495. PMLR.
\newblock ISSN: 2640-3498.

\bibitem[Haarnoja et~al., 2018]{haarnoja_soft_2018}
Haarnoja, T., Zhou, A., Abbeel, P., and Levine, S. (2018).
\newblock Soft {Actor}-{Critic}: {Off}-{Policy} {Maximum} {Entropy} {Deep} {Reinforcement} {Learning} with a {Stochastic} {Actor}.
\newblock arXiv:1801.01290 [cs].

\bibitem[Hekmatmanesh et~al., 2021]{stress}
Hekmatmanesh, A., Zhidchenko, V., Kauranen, K., Siitonen, K., Handroos, H., Soutukorva, S., and Kilpeläinen, A. (2021).
\newblock Biosignals in human factors research for heavy equipment operators: A review of available methods and their feasibility in laboratory and ambulatory studies.
\newblock {\em IEEE Access}, 9:97466--97482.

\bibitem[Huang et~al., 2023]{huangRoboticArmVelocity2023}
Huang, H.-H., Cheng, C.-K., Chen, Y.-H., and Tsai, H.-Y. (2023).
\newblock The {Robotic} {Arm} {Velocity} {Planning} {Based} on {Reinforcement} {Learning}.
\newblock {\em International Journal of Precision Engineering and Manufacturing}, 24(9):1707--1721.

\bibitem[Huh et~al., 2023]{huhDeepLearningBasedAutonomous2023}
Huh, J., Bae, J., Lee, D., Kwak, J., Moon, C., Im, C., Ko, Y., Kang, T.~K., and Hong, D. (2023).
\newblock Deep learning-based autonomous excavation: A bucket-trajectory planning algorithm.
\newblock {\em IEEE Access}, 11:38047--38060.

\bibitem[Janner et~al., 2021]{janner2021offline}
Janner, M., Li, Q., and Levine, S. (2021).
\newblock Offline reinforcement learning as one big sequence modeling problem.
\newblock {\em Advances in neural information processing systems}, 34:1273--1286.

\bibitem[Jash et~al., 2024]{rlAV}
Jash, K.~G., Patel, J., Gupta, R., Jadav, N.~K., Tanwar, S., and Desai, S. (2024).
\newblock Reinforcement learning-based optimized driving behaviour framework for autonomous vehicles in intelligent transportation system.
\newblock In {\em 2024 International Conference on Electrical Electronics and Computing Technologies (ICEECT)}, volume~1, pages 1--6.

\bibitem[Kurinov et~al., 2020]{kurinovAutomatedExcavatorBased2020}
Kurinov, I., Orzechowski, G., Hämäläinen, P., and Mikkola, A. (2020).
\newblock Automated {Excavator} {Based} on {Reinforcement} {Learning} and {Multibody} {System} {Dynamics}.
\newblock {\em IEEE Access}, 8:213998--214006.
\newblock Conference Name: IEEE Access.

\bibitem[La~Hera et~al., 2024]{lahera}
La~Hera, P., Mendoza-Trejo, O., Lindroos, O., Lideskog, H., Lindbäck, T., Latif, S., Li, S., and Karlberg, M. (2024).
\newblock Exploring the feasibility of autonomous forestry operations: Results from the first experimental unmanned machine.
\newblock {\em Journal of Field Robotics}, 41(4):942--965.

\bibitem[Makoviychuk et~al., 2021]{makoviychuk_isaac_2021}
Makoviychuk, V., Wawrzyniak, L., Guo, Y., Lu, M., Storey, K., Macklin, M., Hoeller, D., Rudin, N., Allshire, A., Handa, A., and State, G. (2021).
\newblock Isaac {Gym}: {High} {Performance} {GPU}-{Based} {Physics} {Simulation} {For} {Robot} {Learning}.
\newblock arXiv:2108.10470 [cs].

\bibitem[Mnih et~al., 2016]{mnih_asynchronous_2016}
Mnih, V., Badia, A.~P., Mirza, M., Graves, A., Lillicrap, T.~P., Harley, T., Silver, D., and Kavukcuoglu, K. (2016).
\newblock Asynchronous {Methods} for {Deep} {Reinforcement} {Learning}.
\newblock arXiv:1602.01783 [cs].

\bibitem[Mnih et~al., 2013]{mnih_playing_2013}
Mnih, V., Kavukcuoglu, K., Silver, D., Graves, A., Antonoglou, I., Wierstra, D., and Riedmiller, M. (2013).
\newblock Playing {Atari} with {Deep} {Reinforcement} {Learning}.
\newblock arXiv:1312.5602 [cs].

\bibitem[Narvekar et~al., 2020]{narvekar_curriculum_2020}
Narvekar, S., Peng, B., Leonetti, M., Sinapov, J., Taylor, M.~E., and Stone, P. (2020).
\newblock Curriculum {Learning} for {Reinforcement} {Learning} {Domains}: {A} {Framework} and {Survey}.
\newblock {\em Journal of Machine Learning Research}, 21(181):1--50.

\bibitem[Schulman et~al., 2017]{schulmanProximalPolicyOptimization2017}
Schulman, J., Wolski, F., Dhariwal, P., Radford, A., and Klimov, O. (2017).
\newblock Proximal {Policy} {Optimization} {Algorithms}.
\newblock arXiv:1707.06347 [cs].

\bibitem[Shevchuk et~al., 2021]{shevchuk}
Shevchuk, D., Malysheva, I., Alizadeh, M., and Handroos, H. (2021).
\newblock Simulated and experimental analysis of a log crane with conventional and direct driven hydraulics.
\newblock volume ASME/BATH 2021 Symposium on Fluid Power and Motion Control of {\em Fluid Power Systems Technology}, page V001T01A044.

\bibitem[Strandgard et~al., 2017]{fwdProductivity}
Strandgard, M., Mitchell, R., and Acuna, M. (2017).
\newblock Time consumption and productivity of a forwarder operating on a slope in a cut-to-length harvest system in a pinus radiata d. don pine plantation.
\newblock {\em Journal of Forest Science}, 63:324--330.

\bibitem[Sutton and Barto, 2018]{suttonReinforcementLearningIntroduction2018}
Sutton, R.~S. and Barto, A.~G. (2018).
\newblock {\em Reinforcement {Learning}: {An} {Introduction}}.
\newblock A Bradford Book, Cambridge, Massachusetts, second edition edition.

\bibitem[Tobias et~al., 2024]{semberg}
Tobias, S., Anders, N., Rolf, B., and Linnea, H. (2024).
\newblock Real-time target point identification and automated log grasping by a forwarder, using a single stereo camera for both object detection and boom-tip control.
\newblock {\em Silva Fennica}, 58(1).

\bibitem[Tufano et~al., 2022]{rlGameTesting}
Tufano, R., Scalabrino, S., Pascarella, L., Aghajani, E., Oliveto, R., and Bavota, G. (2022).
\newblock Using reinforcement learning for load testing of video games.
\newblock In {\em 2022 IEEE/ACM 44th International Conference on Software Engineering (ICSE)}, pages 2303--2314.

\bibitem[Wallin et~al., 2024]{wallin}
Wallin, E., Wiberg, V., and Servin, M. (2024).
\newblock Multi-log grasping using reinforcement learning and virtual visual servoing.
\newblock {\em Robotics}, 13(1).

\bibitem[Wiberg et~al., 2022]{wiberg}
Wiberg, V., Wallin, E., Nordfjell, T., and Servin, M. (2022).
\newblock Control of rough terrain vehicles using deep reinforcement learning.
\newblock {\em IEEE Robotics and Automation Letters}, 7(1):390--397.

\bibitem[Zhidchenko et~al., 2025]{hap}
Zhidchenko, V., Startcev, E., and Handroos, H. (2025).
\newblock Blindfolded operation as a method of haptic feedback design for mobile machinery.
\newblock In Kajimoto, H., Lopes, P., Pacchierotti, C., Basdogan, C., Gori, M., Lemaire-Semail, B., and Marchal, M., editors, {\em Haptics: Understanding Touch; Technology and Systems; Applications and Interaction}, pages 323--337, Cham. Springer Nature Switzerland.

\end{thebibliography}
\footnotetext{Images taken from YouTube video: \url{https://www.youtube.com/watch?v=l4uo_z5Hy9w}}
\stepcounter{footnote}
\footnotetext{Code can be accessed from: \url{https://github.com/iku-work/ForwarderOmniIsaacGymEnvs.git}}
\end{document}